\documentclass[11pt]{article}

\usepackage[final]{acl}
\usepackage{times}
\usepackage{latexsym}

\usepackage{array}
\usepackage{ragged2e}
\usepackage{booktabs}
\usepackage{multirow}
\usepackage{graphicx}
\usepackage{tabularx}
\usepackage{amsmath}
\usepackage{placeins}
\usepackage{microtype}
\usepackage{inconsolata}

\usepackage{fontspec}
\usepackage{polyglossia}
\setmainlanguage{english}
\setotherlanguage{arabic}
\newfontfamily\arabicfont[
  Path = ./fonts/,
  Extension = .ttf,
  Script=Arabic,
  Scale=MatchLowercase
]{Amiri-Regular}

\newcolumntype{L}[1]{>{\RaggedLeft\arraybackslash}p{#1}}
\newcolumntype{Y}{>{\raggedright\arraybackslash}X}

\title{BARRAC: Adaptation of an English Aspect-based Sentiment Analysis Approach for Classification Tasks in Arabic Dialects}

\author{
Ali Almutairi$^{1}$,
Gelareh Mohammadi$^{1}$, \\
\textbf{Imran Razzak}$^{2}$,
\textbf{Aditya Joshi}$^{1}$, \\
$^{1}$University of New South Wales, Australia,
$^{2}$MBZUAI, UAE \\
\texttt{ali.almutairi@unsw.edu.au} \\
}

\begin{document}
\maketitle
\begin{abstract}
% With growing NLP research and particularly for Arabic, several models, datasets and benchmarks have been reported for Arabic and its varied dialects. This paper asks if past approaches reported for certain tasks in majority languages like English can be adapted to other tasks for Arabic. We show how a past approach to aspect-based sentiment analysis can be adapted to classification tasks such as sarcasm, dialect, and sentiment. We present the adaptation as BARRAC: Brainstorming Alignment and Replaced Representation learning for ArabiC tasks. We adapt DS²-ABSA by replacing its consumer-review attribute pools and synthesis logic with task-specific Arabic linguistic markers for dialectal sentiment, sarcasm, and dialect classification across a ten-domain space, while replacing noisy self-training with intermediate training on synthetic data followed by few-label fine-tuning. BARRAC framework achieves a mean macro-F1 of 63.93 across five Arabic dialect datasets, outperforming the best supervised baseline by +8.48 points, the best few-label SOTA method by +9.00 points, and matching GPT-4o (63.98) at a fraction of its parameter count while surpassing it on four of five datasets. Error analysis suggests that remaining errors are more closely associated with dataset characteristics, with challenges arising from ambiguous sentiment boundaries, precision--recall trade-offs in sarcasm detection, and confusion among low-support dialect classes. This paper demonstrates how adaptation of approaches is valuable alongside adaptation of models, datasets and benchmarks.
With the rapid growth of Arabic NLP, several models, datasets and benchmarks have been reported. This paper asks whether approaches developed for majority languages like English can be adapted to Arabic tasks. We adapt an English aspect-based sentiment analysis framework to Arabic classification tasks and present the adaptation as BARRAC: Brainstorming Alignment and Replaced Representation learning for ArabiC tasks. BARRAC replaces consumer-review attribute pools with Arabic linguistic devices and markers for dialectal sentiment, sarcasm, and dialect identification, and replaces noisy self-training with two-stage training. Evaluated on five Arabic dialect datasets, BARRAC achieves a mean macro-F1 of 63.93\%, outperforming the best few-label SOTA by 3\%, and outperforming GPT-4o on four out of five tasks. Error analysis provides insights into remaining challenges. These results demonstrate that adapting task-specific approaches is a promising direction for Arabic NLP alongside adapting models, datasets and benchmarks. 
% The code and resources are available at https://github.com/nlp-research-artifacts/BARRAC/.
\end{abstract}

\section{Introduction}

Arabic is a Semitic language with over 1,400 years of written history \cite{intro-Arabic_stats1-belinkov2018} and one of the six official languages of the United Nations, with more than 422 million native speakers \cite{intro-Arabic_stats2-udadNaaima2018SaiA}. Since the early finite-state morphology of \citet{intro-FirstArabicNLP-beesley-1996}, Arabic NLP has produced dedicated pre-trained models \cite{AraBERTv2-Exp-compMod-p1-antoun-etal-2020-arabert, Intro-ARBERTAnd-MARBERT-ARLUE-abdul-mageed-etal-2021}, evaluation suites \cite{intro-alue-seelawi-etal-2021}, recurring NADI dialect-identification shared tasks \cite{intro-nadi-talafha-etal-2025, intro-nadi-abdul-mageed-etal-2024, intro-nadi-abdul-mageed-etal-2023}, and community datasets for sentiment and sarcasm \cite{Ar-Sarcasm-abu-farha-magdy-2020-arabic-Exp-dat-p1, almazrua-etal-2022-sa7r-Exp-dat-p1}. Yet these resources remain limited in scale and diversity compared to English, especially for low-resource tasks such as fine-grained sarcasm detection and cross-dialect sentiment classification \citep{intro-low_resource-abdhood2025}. We argue this gap calls for a community-involved effort in which native speakers critically adapt and refine approaches first devised for English.

We show how DS²-ABSA~\cite{motiva-ds2-absa}, an English aspect-based sentiment analysis framework, can be adapted to text classification tasks in Arabic, including low-resource dialects. Our adaptation is called BARRAC: \textbf{B}rainstorming \textbf{A}lignment and \textbf{R}eplaced \textbf{R}epresentation learning for \textbf{A}rabi\textbf{C} tasks. Existing responses to label scarcity, zero-shot transfer, SFT, and few-label methods, such as Cluster\&Tune \cite{shnarch-etal-2022-cluster}, IDoFew \cite{IDoFew}, and noisy self-training \cite{motiva-ds2-absa}, have not, to our knowledge, extended a complete English data-synthesis framework to Arabic sociolinguistic tasks. Benchmarking 14 baselines across dialectal sarcasm, sentiment, and dialect tasks with few-labels, our adaptation reaches a mean Macro-F1 of 63.93\%, surpassing the best supervised baseline by 8.48\%, SOTA few-label method by 3\% and is comparable to GPT-4o at a fraction of its parameter count, with consistent gains across all five datasets. Methodological contributions devised for English, we conclude, can be meaningfully adapted to serve other languages and communities.

\section{Motivation}
Arabic NLP has grown rapidly, with dedicated venues such as the ArabicNLP conference reflecting the field's increasing maturity \cite{intro-low_resource-abdhood2025}. While acknowledging the focus on building datasets and benchmarks is valuable, we ask an alternative question: ``\textit{how can an English aspect-based sentiment analysis approach be adapted for classification tasks in Arabic?}'' DS²-ABSA is a dual-stream (key-point-driven and instance-driven) synthesis framework addressing label scarcity in Aspect-Based Sentiment Analysis (ABSA). However, its attribute pools are grounded exclusively in consumer-review domains, and it presupposes tangible review subjects with discrete, reviewable aspects such as food quality, making it non-trivially applicable to tasks with a fundamentally different label space, output structure, and subject matter. Via our proposed BARRAC, we demonstrate how DS²-ABSA can be systematically adapted beyond its original scope: from tasks centred on tangible consumer entities to tasks centred on linguistic and sociolinguistic phenomena, specifically, sarcasm, dialects, and sentiments.

% \citet{motiva-ds2-absa} propose DS²-ABSA, a dual-stream data synthesis framework designed to address the scarcity of labelled data for Aspect-Based Sentiment Analysis (ABSA) by leveraging large language models to generate higher-quality synthetic training instances. The framework has two complementary streams: a key-point-driven stream and an instance-driven stream. However, it faces two key constraints. First, its attribute pools are grounded exclusively in consumer review domains, specifically restaurants and laptops, where opinion targets are concrete, nameable physical entities or product features. Second, the framework was designed for tangible review subjects: entities that possess discrete, reviewable aspects such as food quality, ambience, or screen resolution. These design choices make the framework non-trivially applicable to tasks with a fundamentally different label space, output structure, and subject matter. 
% Naively substituting a different dataset while retaining the original attribute definitions and task instruction would produce prompts that are semantically incoherent for single-label classification tasks, where there are no aspect terms to extract, no multi-aspect polarity structure to label, and crucially, no physical entity being reviewed.

\begin{figure}[!t]               
    \centering
    \includegraphics[
        trim=0cm 1.3cm 0cm 0cm,
        clip,
    width=\columnwidth  ]{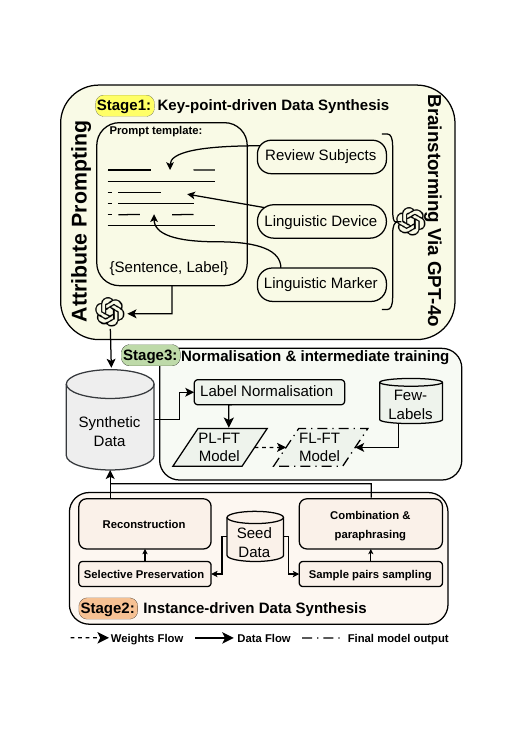}

\captionof{figure}{BARRAC framework.  
}
\label{fig:BARRAC-methodology-diagram}
\end{figure}

\FloatBarrier

\section{Methodology}
\subsection{ABSA Method}

Given a consumer-review sentence, DS\textsuperscript{2}-ABSA outputs one or more (aspect term, sentiment polarity) pairs over two domains, restaurant and laptop reviews. It addresses label scarcity by synthesising training data through two complementary streams. The key-point-driven stream builds samples from scratch: it samples an attribute tuple (review subject, aspect category, aspect term, opinion term), prompts an LLM to generate a sentence satisfying all four, and derives the label from the output. The instance-driven stream diversifies the seed corpus by combining, paraphrasing, and selectively reconstructing masked samples. A final refinement stage applies rule-based normalisation and noisy self-training, where a teacher model trained on the synthetic data re-annotates noisy instances.

\subsection{Adaptation to BARRAC}
% 1-2: We will now go over each of the components in ABSA that we changed.
% Subsection: Each component that is changed
% 1 sentence about what it does.
% 1 sentence about why it is insufficient.
% 1-2 sentences about how it was changed.

% ^^^ Repeat for each component that is changed.
BARRAC (see Figure~\ref{fig:BARRAC-methodology-diagram}) extends the components of DS²-ABSA as follows.

\subsubsection{Attribute Linguistic Device}
% The aspect category is a coarse-grained dimension of a product or service (e.g., screen resolution), brainstormed per domain via GPT-4o. Such categories are tied to physical properties of reviewable entities and have no counterpart when the subject is a linguistic or sociolinguistic phenomenon. We substitute with linguistic devices, respectively: sarcasm devices, coarse rhetorical mechanisms (e.g., mock praise); dialect regions such as Egyptian and Gulf countries, conditioning the hierarchy on the target class directly; and sentiment sub-categories, domain-specific manifestations of each polarity (e.g., frustration with price hikes for negative sentiment in the economy domain). Each preserves the coarse conditioning role from which finer attributes are derived, grounded in the relevant phenomenon rather than product features.

A linguistic device is a coarse-grained rhetorical or communicative mechanism through which a speaker conveys meaning beyond the literal content of an utterance. In the original framework, the aspect category serves an analogous role as a coarse-grained dimension of a product or service (e.g., screen resolution), brainstormed per domain via GPT-4o. Such categories are inherently tied to the physical properties of reviewable entities and have no natural counterpart when the subject of analysis is a linguistic or sociolinguistic phenomenon rather than a product. We substitute aspect categories with linguistic devices conditioned on the target task: for sarcasm detection, we use coarse-grained rhetorical mechanisms such as mock\_praise and verbal\_irony; for dialect identification, we use dialect regions (Egyptian, Gulf, Levantine, Maghrebi, and Modern Standard Arabic), conditioning the attribute hierarchy on the target class directly; and for sentiment analysis, we use sentiment sub-categories representing domain-specific manifestations of each polarity class, such as frustration with price hikes for negative sentiment in the economy domain.
In all three cases, the substituted attribute preserves the coarse conditioning role of the original aspect category, from which finer-grained attributes are subsequently derived, while grounding the hierarchy in the relevant linguistic phenomenon rather than product features. Please refer to Table~\ref{tab:barrac-brainstorm} for further details.

\begin{table*}[t]
\centering
\small
\setlength{\tabcolsep}{4pt}
\begin{tabular}{llrrrrrrr}
\toprule
\textbf{Dataset} & \textbf{Split} &
\textbf{Total} & \textbf{Classes} &
\textbf{Tok. Total} &
\textbf{Tok. Mean} &
\textbf{Tok. Med.} &
\textbf{Tok. Min} &
\textbf{Tok. Max} \\
\midrule

\multirow{4}{*}{DART}
& Train         & --    & -- & --     & --    & --    & -- & -- \\
& Real Labels   & 100   & 5  & 1,447  & 14.47 & 14.00 & 3  & 28 \\
& Eval          & 255   & 5  & 3,645  & 14.29 & 13.00 & 1  & 31 \\
& Test          & 1,009 & 5  & 14,304 & 14.18 & 13.00 & 2  & 35 \\
& Synthetic Labels & 42929 & 5 & 1002785 & 23.36 & 24.00 & 2 & 59 \\
\midrule

\multirow{4}{*}{Ar-Dialects}
& Train         & 5,905 & 5 & 86,630 & 14.67 & 15.00 & 1 & 32 \\
& Real Labels   & 100   & 5 & 1,482  & 14.82 & 15.00 & 3 & 28 \\
& Eval          & 2,432 & 5 & 36,107 & 14.85 & 15.00 & 1 & 29 \\
& Test          & 2,110 & 5 & 31,273 & 14.82 & 15.00 & 1 & 29 \\
& Synthetic Labels & 42929 & 5 & 1002785 & 23.36 & 24.00 & 2 & 59 \\
\midrule

\multirow{4}{*}{Sa'7r}
& Train         & 3,958 & 2 & 57,333 & 14.49 & 10.00 & 1 & 59 \\
& Real Labels   & 100   & 2 & 1,487  & 14.87 & 10.00 & 2 & 54 \\
& Eval          & 3,458 & 2 & 49,352 & 14.27 & 10.00 & 1 & 59 \\
& Test          & 1,186 & 2 & 16,976 & 14.31 & 10.00 & 2 & 59 \\
& Synthetic Labels & 38542 & 2 & 770288 & 19.99 & 21.00 & 2 & 62 \\
\midrule

\multirow{4}{*}{Ar-Sarcasm}
& Train         & 5,905 & 2 & 86,811 & 14.70 & 15.00 & 1 & 29 \\
& Real Labels   & 100   & 2 & 1,421  & 14.21 & 14.00 & 2 & 25 \\
& Eval          & 2,432 & 2 & 35,987 & 14.80 & 15.00 & 1 & 32 \\
& Test          & 2,110 & 2 & 31,273 & 14.82 & 15.00 & 1 & 29 \\
& Synthetic Labels & 38556 & 2 & 857920 & 22.25 & 23.00 & 2 & 53 \\
\midrule

\multirow{4}{*}{Ar-Sentiment}
& Train         & 5,835 & 3 & 85,989 & 14.74 & 15.00 & 1 & 32 \\
& Real Labels   & 100   & 3 & 1,384  & 13.84 & 13.50 & 2 & 29 \\
& Eval          & 2,502 & 3 & 36,846 & 14.73 & 15.00 & 1 & 29 \\
& Test          & 2,110 & 3 & 31,273 & 14.82 & 15.00 & 1 & 29 \\
& Synthetic Labels & 48599 & 3 & 1039146 & 21.38 & 22.00 & 2 & 67 \\

\bottomrule
\end{tabular}
\caption{Statistics of the train, real labels, evaluation, and test splits used in the five Arabic classification datasets. Tok. Total, Tok. Mean, Tok. Med., Tok. Min, and Tok. Max denote the total, mean, median, minimum, and maximum number of tokens per split, respectively.}
\label{tab:dataset_split_statistics}
\end{table*}

\begin{table*}[t]
\centering
\resizebox{\textwidth}{!}{
\begin{tabular}{lcccccc}
\toprule
\textbf{Method/Dataset} &
\textbf{Ar-Sent.} &
\textbf{Ar-Sarc.} &
\textbf{Sa'7r} &
\textbf{DART} &
\textbf{Ar-Dial.} &
\textbf{Mean} \\
\midrule

\multicolumn{7}{c}{\textbf{Without Fine-Tuning}} \\
\midrule
ArabicBERT & 3.27 & 24.68 & 45.77 & 32.63 & 5.30 & 2.72 \\
ARBERTv2 & 2.18 & 11.41 & 47.56 & 32.92 & 10.77 & 7.88 \\
AraBERT-Twitter & 4.14 & 25.02 & 46.76 & 38.52 & 7.40 & 10.14 \\
MARBERTv2 & 5.05 & 20.17 & 50.40 & 51.74 & 10.77 & 6.03 \\

\midrule
\multicolumn{7}{c}{\textbf{SFT with 100 Labels}} \\
\midrule
BERT & 45.72 & 49.34 & 59.91 & 10.12 & 16.02 & 36.22 \\
DistilBERT & 43.74 & 49.44 & 64.90 & 6.98 & 16.02 & 36.22 \\
ArabicBERT & 57.77 & 58.71 & 64.48 & 25.56 & 26.33 & 46.57 \\
ARBERTv2 & 60.25 & 61.91 & 61.88 & 52.67 & 40.58 & 55.46 \\

\midrule
\multicolumn{7}{c}{\textbf{SOTA Few-label Methods with 100 Labels}} \\
\midrule
Cluster\&Tune$^{\ast}$  & 57.98 & 57.05 & 64.90 & 49.75 & 39.79 & 53.89 \\
IDoFew$^{\ast}$  & 54.22 & 60.82 & 63.54 & 53.24 & 32.38 & 52.84 \\
IDoFew$^{\ast}$+  & 58.73 & 61.74 & 66.88 & 51.60 & 35.73 & 54.93 \\

% DS2-ABSA$_{\text{T5-base}}$ & 46.01 & 52.07 & 52.06 & 24.62 & 22.45 & 39.44 \\
% Noisy self-training$^{\ast}$ & 65.31 & 66.43 & 64.20 & 66.43 & 42.21 & 60.92 \\
% Noisy self-training$^{\ast}$+ & 64.06 & 62.07 & 64.24 & 67.82 & 44.37 & 60.51 \\

\midrule
\textbf{BARRAC (Ours)} &
\textbf{65.61} &
\textbf{69.06} &
\textbf{67.03} &
\textbf{68.76} &
\textbf{49.19} &
\textbf{63.93} \\

\bottomrule
\end{tabular}
}
\caption{Mac-F1 results on 5 Arabic datasets. ``$\ast$'' indicates ARBERTv2 as SOTA  backbone model and ``+'' indicates it is adapted to our training hyperparameters.}
\label{tab:sft_100_labels}
\end{table*}

\subsubsection{Attribute Linguistic Marker}
% The aspect term is a fine-grained English noun phrase naming the opinion target (e.g., decor), conditioned on the aspect category. Like categories, it presupposes a discrete physical entity, absent in linguistic classification. We substitute class-conditioned linguistic markers: Arabic phrases signalling the active sarcasm device (e.g., \textarabic{<\textbf{يا سلام}>} for mock praise); phonological, morphological, and lexical features of each dialect region (e.g., \textarabic{<\textbf{إيه ده}>} for Egyptian); and Arabic trigger phrases signalling a polarity (e.g., \textarabic{<\textbf{غلاء فاحش}>} for negative). Markers are brainstormed on the class-level Aspect Term attribute, preserving the coarse-to-fine hierarchy.
A linguistic marker is a fine-grained lexical or phrasal unit
whose presence in an utterance reliably signals the activation of a particular linguistic device, functioning as a surface-level cue from which deeper communicative intent can be inferred. In the original framework, the aspect term serves this role as a fine-grained noun phrase naming the specific opinion target (e.g., decor), conditioned on the aspect category. Like aspect categories, aspect terms presuppose the existence of a discrete, nameable physical entity, which is absent in linguistic classification tasks.
We substitute aspect terms with class-conditioned linguistic markers: for sarcasm, these are Arabic words and phrases that signal the active sarcasm device (e.g., \textarabic{<يا سلام> } for mock\_praise); for dialect identification, these are phonological, morphological, and lexical features characteristic of each dialect region (e.g., \textarabic{<إيه ده> } for Egyptian); and for sentiment analysis, these are Arabic trigger phrases whose presence strongly signals a particular polarity (e.g., \textarabic{<غلاء فاحش> } for negative sentiment). In all three cases, markers are brainstormed conditioned on the class-level linguistic device, preserving the coarse-to-fine hierarchy of the original framework. Please refer to Table~\ref{tab:barrac-brainstorm} for further details.

% \paragraph{Attribute Hierarchy}
% ABSA conditions generation on a two-level attribute hierarchy: a coarse Aspect Category (AC), a dimension of a product or service (e.g., screen resolution), and a fine-grained Aspect Term (AT), an English noun phrase naming the opinion target (e.g., decor), brainstormed conditioned on the category. Both presuppose a discrete physical entity being reviewed, which is absent when the subject is a linguistic or sociolinguistic phenomenon. We preserve the coarse-to-fine structure but reground both levels per task. For sarcasm: sarcasm devices (e.g., mock praise) refined into Arabic phrases (linguistic marker) signalling the active device (e.g., \textarabic{<\textbf{يا سلام}>}). For dialect: dialect regions (EGY, GLF, LEV, MGH) refined into characteristic phonological, morphological, and lexical features (e.g., \textarabic{<\textbf{إيه ده}>} for Egyptian). For sentiment: domain-specific sub-categories of each polarity (e.g., frustration with price hikes for negative sentiment in the economy domain) refined into Arabic trigger phrases (e.g., \textarabic{<\textbf{غلاء فاحش}>}).

\paragraph{Task Instruction}
The original instruction asks the model to ``label the sentence by extracting the aspect term(s) and identifying their corresponding sentiment polarity''. This is not suitable for single-label (predicting one class at a time) tasks, which have no multi-pair structure. We substitute a task-specific objective such as ``identifying whether the text is sarcastic or not sarcastic''. The prompt structure ``Label the [sentence/text] by {task instruction}'' is preserved verbatim, with only the objective replaced in the attribute prompting. Additional details are provided in Table~\ref{tab:barrac-instructions} for task instruction and Table~\ref{tab:barrac-attribute} for attribute prompting.

% \begin{table}[b]
% \footnotesize
% \begin{tabular}{lrrrr}
% \hline 
% \textbf{Dataset} & \textbf{Total} & \textbf{$C$}  & \textbf{Task} \\
% \midrule
% DART         & 1,364  & 5  & Dial. \\
% Ar-Dialects  & 10,547 & 5  & Dial. \\
% Sa'7r        & 8,702  & 2  & Sarc. \\
% Ar-Sarcasm   & 10,547 & 2  & Sarc. \\
% Ar-Sentiment & 10,547 & 3  & Sent.\\
% \bottomrule
% \end{tabular}
% \caption{  ``$C$'' = classes.}
% \label{tab:dataset_overall_statistics}
% \end{table}

\paragraph{Domain Scope}
DS²-ABSA restricts brainstorming to two domains, restaurants and laptops, sourcing both the topic pool and category hierarchy. This suits ABSA but limits diversity and is unsuitable for tasks spanning public discourse. We replace it with 10 domains: economy, sports, health, environment, entertainment, transportation, government services and infrastructure, technology and the internet, education, and food.

Examples of Steps 1 and 2 in BARRAC are provided in Tables~\ref{tab:barrac-example-kpd} and~\ref{tab:barrac-example-idd}, respectively.

\paragraph{Training Strategy}
Instead of DS²-ABSA's noisy self-training, which risks error accumulation across iterative re-annotation~\citep{zou-caragea-2023-jointmatch}, we employ intermediate training~\citep{IDoFew}: The Pseudo-Label Fine-Tuning (PL-FT) model is trained using pseudo-labels, which will be obtained from systematic synthesis data pipelines. Once training is complete, we utilise the weights from the PL-FT model as initialisation weights for the subsequent Few-Labels Fine-Tuning (FT-FL) model on actual labelled data (see Figure~\ref{fig:BARRAC-methodology-diagram} Stage 3). The synthetic corpus acts as task-adaptive initialisation, while the final decision boundary is refined solely through gold-standard supervision.

% Add main results table here

\section{Experiment Setup}
% 1 sentence describing the tasks and dataset sizes.
We evaluate our adaptation to DS2-ABSA for text classification with few-labels across 5 publicly available Arabic datasets: Ar-Sentiment, Ar-Sarcasm, Ar-Dialects \cite{Ar-Sarcasm-abu-farha-magdy-2020-arabic-Exp-dat-p1}, Sa'7r \cite{almazrua-etal-2022-sa7r-Exp-dat-p1}, and DART \cite{alsarsour-etal-2018-dart-Exp-dat-p1}; spanning three tasks, sarcasm detection, dialect identification, and sentiment analysis.  We fine-tune all models using 100 real labels to mimic low-resource settings. For synthetic data generation, we follow the hyperparameters reported in \citet{motiva-ds2-absa} without modification, except for the LLM (GPT-4o) used for generation. 
% See Appendix~\ref{sec:para_data} (Tables~\ref{tab:hyperparams},~\ref{tab:dataset_split_statistics}) for real and synthetic datasets and setup details.

 We release the synthetic data used in our experiments,
% \footnote{\url{https://huggingface.co/datasets/Ali-a11/BARRAC-datasets}}.
% 1 sentence about base models used within the adapted version.
We use ARBERTv2 as the backbone classification model across all experiments for the few-label frameworks. 
% 1-2 sentences about baselines.

% 1 -2 sentences about the metrics used.
We report the macro-F1 score, consistent with prior work on Arabic text classification under class imbalance. For all baselines, we use a unified configuration: the same number of epochs, batch size, learning rate, optimiser, and random seed across all experiments. All baselines are trained using exactly 100 balanced labelled examples. Table~\ref{tab:hyperparams} summarises the hyperparameters used for intermediate and final training across all three tasks.

\subsection{Comparative Models} \label{app-baseline}
We compare against general-purpose and Arabic-specific pre-trained language models fine-tuned on the available labelled data, namely BERT, DistilBERT, ArabicBERT, and ARBERTv2. We further compare against Cluster\&Tune, IDoFew and its Arabic-adapted variants, as well as the noisy self-training from the DS²-ABSA paper and its variants.
\emph{\textbf{BERT models:}} We have compared BARRAC with publicly available language models that include BERT \cite{BERT-Exp-compMod-p1-devlin-etal-2019-bert} and DistilBERT \cite{DistilBERT-Exp-compMod-p1-sanh2020distilbertdistilledversionbert}.

\noindent \emph{\textbf{Arabic-BERT models:}} We have also used the Arabic BERT in our study. These models are: ArabicBERT \cite{ArabicBERT-Exp-compMod-p1-etal-2020-kuisail}, ARBERTv2 \cite{Intro-ARBERTAnd-MARBERT-ARLUE-abdul-mageed-etal-2021}, AraBERT-Twitter \cite{AraBERTv2-Exp-compMod-p1-antoun-etal-2020-arabert}, and MARBERTv2 \cite{Intro-ARBERTAnd-MARBERT-ARLUE-abdul-mageed-etal-2021}. 

\noindent \emph{\textbf{LLMs:}} We evaluate BARRAC against fine-tuned LLMs through Supervised Fine-Tuning (SFT), as they have shown effectiveness in managing intricate tasks. To achieve this, we select AllAM-7B \cite{res-baseline-allam}, Llama3.1-8B \cite{Res-Comp-p2-Llama-3}, AceGPT \cite{Res-Comp-p2-AceGPT-zhu2024second}, Qwen2.5-72B-instruct \cite{res-baseline-qwen2.5}, and GPT-4o \cite{res-baseline-gpt4.1}  as representative models in our comparison. 

%We extend our comparison to LLMs as it shows their power in complex tasks such as generations and coherence. We pick Llama3-8B \cite{} and AceGPT \cite{}.   

\noindent \emph{\textbf{Intermediate models:}} In order to assess the efficacy of our intermediate stage, we conducted a comparison of BARRAC's performance with that of Cluster \& Tune \cite{shnarch-etal-2022-cluster} and IDoFew \cite{IDoFew}, both of which utilise an inter-training task. 

\begin{table}[t]
\centering
\small
\begin{tabular}{ll}
\toprule
\textbf{Parameter} & \textbf{Value} \\
\midrule
PL-FT model epochs & 1 \\
FL-FT model epochs & 20 \\
Learning rate (all other datasets) & $5\times10^{-5}$ \\
Learning rate (Sa'7r) & $3\times10^{-5}$ \\
Batch size & 32 \\
Optimizer & Adam \\
Seed & 42 \\
\bottomrule
\end{tabular}
\caption{Training hyperparameters used across all datasets and baselines.}
\label{tab:hyperparams}
\end{table}

\section{Results}
\label{sec:results}

\iffalse
\begin{table*}[t]
\centering
\resizebox{0.7\textwidth}{!}{
\begin{tabular}{lcccccc}
\toprule
\textbf{BARRAC Component} &
\textbf{Ar-Sent.} &
\textbf{Ar-Sarc.} &
\textbf{Sa'7r} &
\textbf{DART} &
\textbf{Ar-Dial.} &
\textbf{Mean} \\
\midrule
w/o Key-Point-Driven & 64.08 & 65.16 & 67.55 & 67.62 & 47.18 & 62.32 \\
w/o Instance-Driven  & 59.69 & 64.35 & 62.91 & 66.37 & 45.77 & 59.82 \\
w/o Normalisation    & 63.83 & 65.48 & 65.75 & 67.30 & 47.18 & 61.91 \\
w/o Inter-model (PL-FT)               & 60.25 & 61.91 & 61.88 & 52.67 & 40.58 & 55.46 \\
w/o Final model (FL-FT)              & 16.06 & 60.98 & 59.95 & 32.90 & 13.47 & 36.67 \\
\midrule
\textbf{Full Model (Ours)} &
\textbf{65.61} &
\textbf{69.06} &
\textbf{67.03} &
\textbf{68.76} &
\textbf{49.19} &
\textbf{63.93} \\
\bottomrule
\end{tabular}
}
\caption{
Ablation study results reported using Macro-F1 (\%).
}
\label{tab:ablation_study}
\end{table*}
\fi

% \subsection{Comparison with BBMs and Few-Label SOTA}
Table~\ref{tab:sft_100_labels} compares BARRAC against three baseline groups. Direct inference BERT-based models (BBMs) perform near chance (means of 20.97--27.62), confirming that task supervision is indispensable.
% \footnote{Zero-shot rows are included for completeness; they constitute a lower bound rather than a competitive baseline.} 
Supervised fine-tuning (SFT) on the 100 labels lifts the strongest Arabic encoder, ARBERTv2, to a mean of 55.46. The BERT and DistilBERT collapse on DART (10.12 and 6.98), predicting only a subset of the five dialect classes: their English-centric vocabularies fragment dialect-bearing tokens. The few-label SOTA methods Cluster\&Tune \cite{shnarch-etal-2022-cluster} and IDoFew \cite{IDoFew}, both instantiated with ARBERTv2, do not surpass plain ARBERTv2 SFT on average (53.89 and 54.93), suggesting that intermediate clustering objectives developed for English text classification transfer imperfectly to dialectal Arabic tasks.

BARRAC achieves the best macro-F1 on all five datasets, with a mean of 63.93: +8.48 over the best SFT baseline and +3 over the best few-label SOTA method. The improvements are consistent across binary and multi-label spaces, dataset sizes, and both multi-dialect and Saudi-only distributions, indicating that the dual-stream paradigm generalises well beyond its original ABSA setting. Within BARRAC, backbone and training recipe both matter: running the adapted synthesis pipeline with the original DS²-ABSA configuration, with an Arabic encoder under noisy self-training attains 60.92. Since this comparison varies the backbone and recipe jointly, we isolate the training-strategy effect separately in (\S\ref{sec:strategy}).

% \subsection{Comparison with LLMs}
\label{sec:llm_comparison_sec}
Table~\ref{tab:llm_comparison} compares BARRAC against multiple LLMs. We outperform open-weight LLMs by a wide margin and match GPT-4o on average (63.93 vs.\ 63.98), surpassing it on four of the five datasets. See \ref{app:llm} for the full breakdown.

\paragraph{Zero-seed transfer to DART.}
DART provides our strictest generalisation test: unlike the other four datasets, no DART data was used to seed synthesis; the dialect synthetic corpus was generated from Ar-Dialects seeds, and DART contributes only its 100 gold labels, evaluation, and test sets. Despite this, our method reaches 68.76 on DART, +16.09 over ARBERTv2 SFT and +15.52 over the best SOTA method. The synthetic dialect data thus encodes transferable, class-discriminative dialectal evidence rather than memorising seed-set artefacts, which is precisely the property required of synthetic data in low-resource settings.

\paragraph{Comparison with  LLMs.}
    \label{app:llm}
Table~\ref{tab:llm_comparison} benchmarks against LLMs, selected to span scale (7B--72B and a frontier closed model), language specialisation (Arabic-centric ALLaM and AceGPT vs.\ multilingual Llama and Qwen), and accessibility (open vs.\ closed weights). Our 163M-parameter encoder outperforms every open-weight LLM by a wide margin (best: ALLaM-7B at 56.77, $-7.16$ vs.\ ours); notably, Arabic-centric pre-training alone does not close the gap, and scale does not help (Qwen2.5-72B trails ALLaM-7B by 17.20). Most strikingly, our model is comparable on average to GPT-4o (63.93 vs.\ 63.98, the same model that generated our synthetic data) and surpasses it on four of the five datasets. The student thus matches its teacher at a small fraction of the parameter count and inference cost, and without API dependence. GPT-4o retains a clear advantage on Ar-Sentiment (72.47 vs.\ 65.61).

\subsection{Effect of the Synthetic--Gold Training Strategy}
\label{sec:strategy}
In Table \ref{tab:ds2_training_strategies} we contrast two ways of combining synthetic and gold data on the same ARBERTv2 backbone. \emph{Noisy self-training}, as in the original DS²-ABSA, mixes pseudo-labelled synthetic data with gold data throughout training; we evaluate it under both the original hyper-parameters (60.92) and hyper-parameters tuned to our setting (60.51). Our \emph{intermediate training} strategy, fine-tuning first on synthetic data, then on the 100 gold labels, reaches 63.93, a +3.01 improvement over the best self-training variant, with gains on all datasets, largest on Ar-Dialects (+4.82) and Ar-Sarcasm (+2.63) in a non-iterative learning protocol. Because synthetic labels are noisier than gold labels, interleaving them allows label noise to persist into the final model, whereas intermediate training treats synthesis as task-adaptive initialisation and reserves the final optimisation steps for clean supervision. The training strategy choice is consequential, accounting for roughly a third of our total improvement over direct SFT.

\begin{table}[t]
\centering
\scriptsize
\setlength{\tabcolsep}{0.6pt}
\begin{tabular}{p{0.18\textwidth}lcccccc}
\toprule
\textbf{Model} &
\textbf{Ar-Sent.} &
\textbf{Ar-Sarc.} &
\textbf{Sa'7r} &
\textbf{DART} &
\textbf{Ar-Dial.} &
\textbf{Mean} \\
\midrule

\textbf{Ours$_{\text{Intermediate training}}$}
& \textbf{65.61}
& \textbf{69.06}
& \textbf{67.03}
& \textbf{68.76}
& \textbf{49.19}
& \textbf{63.93} \\

Noisy self-training & 65.31 & 66.43 & 64.20 & 66.43 & 42.21 & 60.92 \\
Noisy self-training+ & 64.06 & 62.07 & 64.24 & 67.82 & 44.37 & 60.51 \\

% DS2-ABSA$_{\text{ARBERTv2}}$ +
% & 64.06 & 62.07 & 64.24 & 67.82 & 44.37 & 60.51 \\

% DS2-ABSA$_{\text{ARBERTv2}}$ 
% & 65.31 & 66.43 & 64.20 & 66.43 & 42.21 & 60.92 \\

% DS2-ABSA$_{\text{T5-base}}$
% & 46.01 & 52.07 & 52.06 & 24.62 & 22.45 & 39.44 \\

\bottomrule
\end{tabular}
\caption{A comparative analysis of our proposed method and the DS2-ABSA training strategies was conducted across five distinct evaluation datasets, maintaining the consistency of the ARBERTv2 backbone model. "+" indicates SOTA baselines adapted to our training hyperparameters.}
\label{tab:ds2_training_strategies}
\end{table}

\begin{table}[t]
\centering
\scriptsize
\setlength{\tabcolsep}{0.6pt}
\begin{tabular}{p{0.18\textwidth}lcccccc}
\toprule
\textbf{Method/Dataset} &
\textbf{Ar-Sent.} &
\textbf{Ar-Sarc.} &
\textbf{Sa'7r} &
\textbf{DART} &
\textbf{Ar-Dial.} &
\textbf{Mean} \\
\midrule

\multicolumn{7}{c}{\textbf{SFT using 100 Labels (LLMs)}} \\
\midrule

ALLaM-7B-Instruct & 63.28 & 67.07 & 61.72 & 54.99 & 36.79 & 56.77 \\
Llama 3.1 8B             & 56.81 & 58.32 & 57.94 & 36.35 & 29.92 & 47.87 \\
AceGPT-16B               & 16.89 & 45.55 & 32.54 & 0.35  & 16.02 & 22.27 \\
Qwen2.5-72B-Instruct     & 53.53 & 56.99 & 34.11 & 27.56 & 25.66 & 39.57 \\
GPT-4o                   & \textbf{72.47} & 67.14 & 65.18 & 66.49 & 48.63 & \textbf{63.98} \\

\midrule

\textbf{BARRAC (Ours)} &
65.61 &
\textbf{69.06} &
\textbf{67.03} &
\textbf{68.76} &
\textbf{49.19} &
63.93 \\

\bottomrule
\end{tabular}
\caption{Macro-F1 (\%) results on against LLMs.}
\label{tab:llm_comparison}
\end{table}

\subsection{Ablation Study}
Table~\ref{tab:ablation_study} ablates BARRAC. On the synthesis side, removing the instance-driven stream costs $-4.11$ mean F1, more than twice the cost of removing the key-point-driven stream $-1.61$; both streams help on nearly every dataset, highlighting their complementarity. Removing label normalisation costs $-2.02$. On the training side, removing the synthetic stage entirely costs $-8.47$. Conversely, removing the gold fine-tuning stage is catastrophic $-27.26$: Ar-Sentiment and Ar-Dialects collapse to 16.06 and 13.47, while binary sarcasm degrades more gracefully (60.98 and 59.95). This asymmetry suggests that synthetic data alone suffices to learn coarse binary distinctions but cannot calibrate fine-grained decision boundaries under the distribution shift between LLM-generated and real tweets; even 100 gold labels are decisive for anchoring the label space. Together, the two training-side ablations show that synthetic and gold data are not interchangeable but strictly complementary: each stage alone underperforms their sequential combination by 8.5 and 27.3 points, respectively.

\section{Error Analysis}
To better understand the remaining failure modes, we manually examined misclassified instances from each dataset. Specifically, we randomly sampled 20 misclassified examples per dataset and grouped them according to the underlying source of error. As shown in Table~\ref{tab:our_error_examples}, most errors fall into four broad categories: Insufficient Context, where very short texts provide limited evidence; Linguistic Overlap, where lexical cues are shared across classes; Label Ambiguity, where multiple interpretations are plausible; and Annotation Ambiguity, where the gold label appears debatable. See Appendix~\ref{sec:error_app} (Figure~\ref{fig:error_pies}, Table~\ref{tab:our_error_examples}) for details.

\begin{table}[t]
\centering
\resizebox{\columnwidth}{!}{
\begin{tabular}{lcccccc}
\toprule
\textbf{BARRAC w/o} &
\textbf{Ar-Sent.} &
\textbf{Ar-Sarc.} &
\textbf{Sa'7r} &
\textbf{DART} &
\textbf{Ar-Dial.} &
\textbf{Mean} \\
\midrule
Key-Point-Driven & 64.08 & 65.16 & 67.55 & 67.62 & 47.18 & 62.32 \\
Instance-Driven  & 59.69 & 64.35 & 62.91 & 66.37 & 45.77 & 59.82 \\
Normalisation    & 63.83 & 65.48 & 65.75 & 67.30 & 47.18 & 61.91 \\
Inter-model (PL-FT)               & 60.25 & 61.91 & 61.88 & 52.67 & 40.58 & 55.46 \\
Final model (FL-FT)              & 16.06 & 60.98 & 59.95 & 32.90 & 13.47 & 36.67 \\
\midrule
\textbf{Full Model} &
\textbf{65.61} &
\textbf{69.06} &
\textbf{67.03} &
\textbf{68.76} &
\textbf{49.19} &
\textbf{63.93} \\
\bottomrule
\end{tabular}
}
\caption{
Ablation study results reported using Macro-F1 (\%).
}
\label{tab:ablation_study}
\end{table}

\section{Conclusion}
We asked whether a task-specific data-synthesis framework devised for English aspect-based sentiment analysis could be adapted to Arabic sociolinguistic classification. By regrounding the components of DS²-ABSA in task-specific Arabic linguistic markers across a ten-domain space, and replacing noisy self-training with intermediate training followed by few-label fine-tuning, our adaptation attains a mean macro-F1 of 63.93 across five datasets and three tasks. In our experiments, it improves on the strongest supervised baseline by 8.48\% and the best few-label method by +3\%, and performs comparably to GPT-4o 63.98 on average while exceeding it on four of five datasets, at a substantially smaller parameter count. These gains hold across binary, three-way, and five-way label spaces and transfer to a zero-seed dataset.

\section{Limitations}
Several factors limit the scope of our conclusions. The generation pipeline itself, however, relies on a single proprietary model (GPT-4o), so regenerating the data from scratch or extending the approach to new tasks incurs API cost, offers limited transparency into the generation process, and may effectively bound the student's performance by the teacher's competence; the gap that remains on Ar-Sentiment would be consistent with such a ceiling, though we cannot establish this from our experiments alone. Our evaluation covers a single language, five datasets, three tasks, and a 100-label budget, so we make no claims about behaviour beyond this particular low-resource setting, and results may not transfer to other languages, dialects, or tasks. The class-balanced synthetic corpus that appears to strengthen minority-class evidence may also shift decision thresholds, which we tentatively associate with the precision cost observed in sarcasm detection; with only 100 gold labels, the fine-tuning stage may be insufficient to fully recalibrate it. Performance on low-support dialects (e.g., Gulf, Levantine, and Maghrebi) is likewise limited, plausibly in part because of an imperfect match between synthetic and real usage. Finally, some of the residual error may not be purely learnable: prior work suggests the neutral--polar boundary in Ar-Sentiment and the irony-hashtag construction of Sa'7r introduce label ambiguity in the source corpora, which could upper-bound achievable performance independently of the model.

\section{Ethical Considerations}
All datasets used are publicly available and were, to our knowledge, employed in accordance with their licenses and intended research use.

\section{Generative AI usage disclosure}
The authors disclose the use of generative AI tools to assist with translation, LaTeX code cleanup, formatting, and the addition of Arabic script.

\bibliography{custom}

\appendix

\section{Extended Error Analysis}
    \label{sec:error_app}
\begin{figure*}[t]
    \centering
    \includegraphics[width=\textwidth]{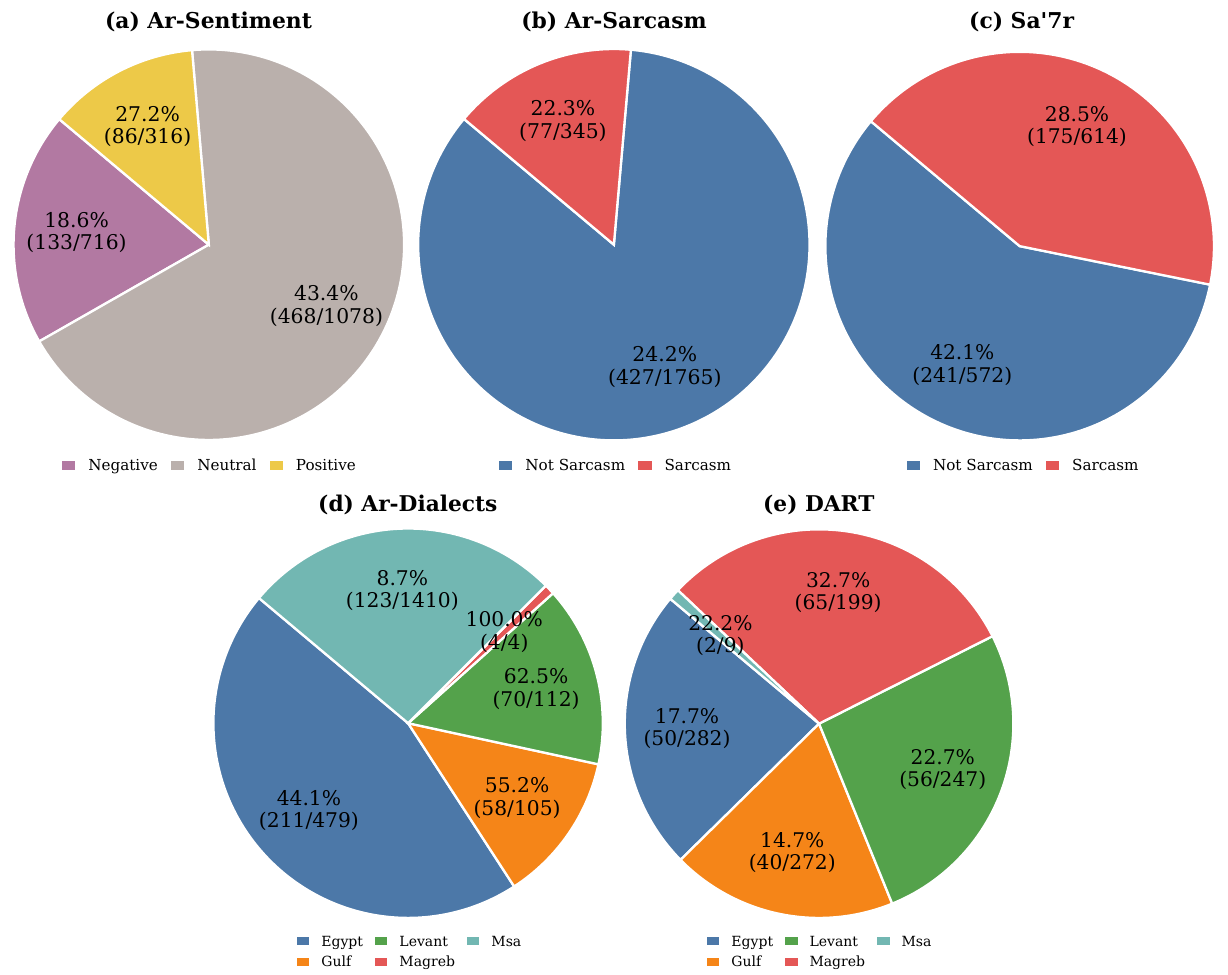}
    \caption{Misclassified instances distribution across five Arabic datasets. Percentages are computed over the test set.}
    \label{fig:error_pies}
\end{figure*}

Figure~\ref{fig:error_pies} reports our model's error rate per gold class (misclassified instances over class support) on each test set; Table~\ref{tab:error_examples} presents representative instances that our model classifies correctly while all competing few-label baselines fail. Examples are drawn from the intersection of baseline errors and our model's correct predictions to characterise the source of the gains in \S\ref{sec:results}; our model's own dominant failure modes, visible in Figure~\ref{fig:error_pies}, are discussed throughout this section.

On Ar-Sentiment, errors are far from uniform across polarities: the neutral class is misclassified at 43.4\%, against 27.2\% for positive and 18.6\% for negative. This mirrors the documented subjectivity of neutral annotations in the source corpus, where re-annotation shifted a large fraction of neutral and positive labels \cite{Ar-Sarcasm-abu-farha-magdy-2020-arabic-Exp-dat-p1}: the neutral--polar boundary is partly an annotation artefact rather than a purely learnable distinction. The qualitative examples reflect the same phenomenon; all baselines mislabel a congratulatory news headline as positive on surface lexis (\textarabic{يهنئ}, ``congratulates''), whereas our model, exposed to synthetic neutral instances containing polar trigger words, correctly returns neutral. Residual neutral errors of this kind plausibly account for much of the remaining gap to GPT-4o on this task (\S\ref{sec:llm_comparison_sec}).

On Ar-Sarcasm, error rates are nearly balanced across classes (24.2\% on not-sarcasm vs.\ 22.3\% on sarcasm), i.e., sarcastic recall reaches 77.7\%, a sharp contrast with the recall bottleneck characteristic of this dataset, whose original baseline detected only 38\% of sarcastic tweets \cite{Ar-Sarcasm-abu-farha-magdy-2020-arabic-Exp-dat-p1}. The cost is precision: because the sarcastic class comprises only 16\% of the test set, the 24.2\% false-alarm rate on the majority class accounts for 427 of the 504 total errors. We attribute this trade-off to our class-balanced synthetic corpus, which strengthens minority-class evidence but shifts the decision threshold toward sarcasm; with only 100 gold labels, the subsequent fine-tuning stage cannot fully re-calibrate it. On Sa'7r the false-alarm tendency is amplified (42.1\% on not-sarcasm vs.\ 28.5\% on sarcasm), which we trace to the corpus construction: Sa'7r was collected via irony-indicative hashtags, keywords, and phrases, so its non-ironic tweets contain the very surface markers that signal irony elsewhere \cite{almazrua-etal-2022-sa7r-Exp-dat-p1}, weakening exactly the marker-level evidence our synthesis injects. The Zamalek example in Table~\ref{tab:error_examples} illustrates the converse skill: all baselines misread enthusiastic, hyperbolic fan discourse as sarcastic, while our model correctly withholds the sarcasm label.

On Ar-Dialects, error rates are inversely related to gold-class support: MSA, with 1,410 test instances, is misclassified at only 8.7\%, while Gulf (n=105) and Levantine (n=112) reach 55.2\% and 62.5\%, and all four Maghrebi instances are missed, though with n=4 this cell measures support scarcity, not model behaviour, and it depresses macro-F1, partly explaining why Ar-Dialects remains our lowest-scoring task (49.19). With 100 gold labels spread over five classes ($\approx$20 per class), the gold stage cannot calibrate rare dialects whose synthetic representations imperfectly match real usage. Zero-seed DART, whose test set is dialect-balanced, shows a strikingly different profile: error rates are uniform and moderate (14.7\%--32.7\%; the 2/9 MSA cell is again too small to interpret), with no class collapsing, further evidence that the synthetic dialect evidence transfers as generalisable markers rather than corpus-specific artefacts (\S\ref{sec:results}). Maghrebi remains the hardest dialect (32.7\%), consistent with DART's own annotation study, which found Maghrebi the least reliably identifiable group even for human annotators \cite{alsarsour-etal-2018-dart-Exp-dat-p1}. The dialect examples in Table~\ref{tab:error_examples} show the characteristic baseline failure our model avoids: collapsing dialectal tweets onto MSA or Levnt, whereas our model recovers the gold dialect from sparse function-word evidence (e.g., Levantine \textarabic{مصاري}, \textarabic{هالكلام}).

\begin{table*}[t]
\centering
\scriptsize
\setlength{\tabcolsep}{4pt}
\renewcommand{\arraystretch}{1.15}
\begin{tabular}{p{0.12\textwidth} p{0.14\textwidth} p{0.48\textwidth} c c}
\toprule
\textbf{Dataset} &
\textbf{Error Type} &
\textbf{Example} &
\textbf{Gold} &
\textbf{Pred.} \\
\midrule

\multirow{5}{*}{%
    \shortstack{\textbf{DART /}\\\textbf{Ar-Dialects}}%
}
& Insufficient Context
& \textarabic{حلو حلو}
  \newline\textit{Nice, nice.}
& MGH & LEV \\

& Linguistic Overlap
& \textarabic{قلت سابقًا وأكرر}
  \newline\textit{I said it before and I repeat it.}
& MGH & MSA \\

& Linguistic Overlap
& \textarabic{اذا هلأ بدّعت}
  \newline\textit{If you excelled now.}
& LEV & GLF \\

& Questionable Annotation
& \textarabic{دخليلك اذا بدهم يجو علي شاكله جيلبرت زوين ونايله معوض بلاها هالقصه}
  \newline\textit{Please, if they are going to come in the style of Gilbert Zwain and Nayla Mouawad, forget about this whole story.}
& MSA & LEV \\

& Label Ambiguity
& \textarabic{ربنا يوفقو ان شاء الله}
  \newline\textit{May God grant him success, God willing.}
& LEV & MGH \\

\midrule

\multirow{6}{*}{%
    \shortstack{\textbf{Ar-Sarcasm /}\\\textbf{Sa'7r}}%
}
& Questionable Annotation
& \textarabic{جبت الكراسه وجبت الالوان عشان حصه الرسم شاطر بس التدريس واصولها ف ميت عقبه وبس ي كومبارس}
  \newline\textit{I brought the notebook and colours for the art class. You're good at that, but teaching and its principles are a hundred obstacles away, you extra.}
& Not-Sarc. & Sarc. \\

& Label Ambiguity
& \textarabic{قل قسم هه}
  \newline\textit{Say ``I swear'', haha.}
& Not-Sarc. & Sarc. \\

& Linguistic Overlap
& \textarabic{لا صراحه هو جاب العيد مع باريس اختياراته سيئه كانت}
  \newline\textit{Honestly, he completely messed it up with Paris; his choices were bad.}
& Not-Sarc. & Sarc. \\

& Questionable Annotation
& \textarabic{الحمدلله الحمدلله اللهم لك الحمد والشكر}
  \newline\textit{Praise be to God, praise be to God; all thanks and gratitude belong to You, O God.}
& Sarc. & Not-Sarc. \\

& Label Ambiguity
& \textarabic{لا يابابا كذا غلط تبي افصل النت عنك}
  \newline\textit{No, baby, that's wrong. Do you want me to disconnect the internet from you?}
& Sarc. & Not-Sarc. \\

& Insufficient Context
& \textarabic{حاجة كدا ع الماشي}
  \newline\textit{Just something quick.}
& Not-Sarc. & Sarc. \\

\midrule

\multirow{5}{*}{\textbf{Sentiment}}
& Questionable Annotation
& \textarabic{البنات اللي م صامو بقولكم ترا رمضان قرب و الوقت قليل ف اللي ناسيين حبيت اذكرهم}
  \newline\textit{Girls who have not started fasting, Ramadan is near and time is short, so I wanted to remind those who forgot.}
& Neu. & Pos. \\

& Label Ambiguity
& \textarabic{الحمد الله وبعد مشوار طويل عن بحث \#بوكيمون جارى التحميل}
  \newline\textit{Thank God, after a long search, Pokémon is finally downloading.}
& Neu. & Pos. \\

& Questionable Annotation
& \textarabic{طقس كاذب يقولو ثلوج ويطلع حر \#طقس\_العرب}
  \newline\textit{False weather forecast; they said there would be snow but it turned out hot.}
& Neu. & Neg. \\

& Insufficient Context
& \textarabic{اى والله :) \#اسلوب\_حياة}
  \newline\textit{Yes indeed :) \#Lifestyle}
& Pos. & Neu. \\

& Questionable Annotation
& \textarabic{ابي كتاب هاري بوتر :( لمن نقلنا بيتنا ذا ضاع الصندوق الي فيه كتبي}
  \newline\textit{I want my Harry Potter book :( When we moved house, the box containing my books got lost.}
& Pos. & Neg. \\

\bottomrule
\end{tabular}
\caption{Representative errors made by our model. Most errors arise from insufficient context, linguistic overlap between classes, label ambiguity, or questionable annotations.}
\label{tab:our_error_examples}
\end{table*}

\newcolumntype{T}[1]{>{\RaggedRight\arraybackslash}p{#1}}
\begin{table*}[t]
\centering
\scriptsize
\setlength{\tabcolsep}{3pt}
\renewcommand{\arraystretch}{1.25}
\begin{tabular}{T{0.45\textwidth}ccccc}
\toprule
\textbf{Text / English Translation} &
\textbf{Gold} &
\textbf{C\&T} &
\textbf{IDoFew} &
\textbf{NST} &
\textbf{Ours} \\
\midrule

\multicolumn{6}{l}{\textbf{Dialect Identification}} \\
\midrule

\textarabic{تعطيك مصاري على هالكلام} \newline
\textit{They give you money for saying this.}
& LEV & GLF & GLF & GLF & \textbf{LEV} \\

\textarabic{مدام أن جنابك شام لجنابي وأنت شمت كثر خيرك وأنا مانب لاجايع ولا ضامي} \newline
\textit{Since you and your side is leaving mine, I'll leave you. May you be blessed with much good; I am neither in need nor asking.}
& GLF & MGH & MGH & MGH & \textbf{GLF} \\

\textarabic{مقاطعة فرنسا المغرب ينظم إلى الاتحاد الإفريقي قريباً ...} \newline
\textit{France boycott, Morocco is joining the African Union soon ...}
& MGH & MSA & MSA & MSA & \textbf{MGH} \\

\textarabic{وزعلانين اذا رفعنا سعر التذاكر .. غصب عنك تدفع وانت ساكت عشان تشوف الاتحاد وجمهوره} \newline
\textit{And they are upset if we raise ticket prices. You will pay whether you like it or not and stay quiet so you can watch Al-Ittihad and its fans.}
& GLF & EGY & EGY & EGY & \textbf{GLF} \\

\midrule
\multicolumn{6}{l}{\textbf{Sarcasm Detection}} \\
\midrule

\textarabic{الحين تدخلون رومات أنتم الحمدلله والشكر} \newline
\textit{Now you enter online chat rooms; thank God.}
& sarcasm & not-sarcasm & not-sarcasm & not-sarcasm & \textbf{sarcasm} \\

\textarabic{عفوا فهد بن محمد اشوف كلامه عن غير اسمها} \newline
\textit{Excuse me, Fahad bin Mohammed's words are about something else.}
& not-sarcasm & sarcasm & sarcasm & sarcasm & \textbf{not-sarcasm} \\

\textarabic{ ﺍﻧﺖ ﻣﻬﻤﺎ ﻛﺒﺮﺕ مش هنسي اي لاعب لبس تيشيرت الزمالك ولعب علشان الزمالك مش علشان الفلوس وحب الزمالك وجمهور الزمالك من قلبه } \newline
\textit{What matters is that any player wears Zamalek's shirt and plays for Zamalek ...}
& not-sarcasm & sarcasm & sarcasm & sarcasm & \textbf{not-sarcasm} \\

\midrule
\multicolumn{6}{l}{\textbf{Sentiment Classification}} \\
\midrule

\textarabic{الاتحاد الأوروبي يهنئ لبنان على تشكيل حكومته الجديدة برئاسة سعد الحريري} \newline
\textit{The European Union congratulates Lebanon on forming its new government headed by Saad Hariri.}
& neutral & positive & positive & positive & \textbf{neutral} \\

\textarabic{طقس العرب أخباركم دائماً متأخرة ولا عندكم مصدر لتحري الطقس يخص بكم} \newline
\textit{Arab Weather, your news is always late, and you do not have your own source for checking the weather.}
& negative & neutral & neutral & positive & \textbf{negative} \\

\bottomrule
\end{tabular}

\caption{
Examples of correctly classified instances by the proposed model where competing baselines make incorrect predictions. Arabic texts are shown with English translations. C\&T denotes Cluster\&Tune, and NST denotes noisy self-training.
}
\label{tab:error_examples}
\end{table*}

\begin{table*}[t]
\centering
\footnotesize
\begin{tabular}{p{2.2cm}|p{12.5cm}}
\hline
\textbf{Attribute} & \textbf{Prompt} \\
\hline

Tweet Topic
& Brainstorm a list of Arabic Twitter tweet topics (at least 200).
\newline\newline
Please adhere to the following guidelines:
\newline
- Summarise each topic as a short, neutral description of a real-world event,
situation, or subject that Arabic Twitter users commonly discuss.
\newline
- Cover diverse domains: economy, sports, health, environment, entertainment, transportation, government services and infrastructure, technology and the internet, education, and food culture.
\newline
- Topics must be neutral, without sentiment, dialect, or rhetorical bias.
\newline\newline
Your output should be a Python list of strings, with each element being a topic
description. Output ONLY the Python list, nothing else. \\
\hline

Linguistic Device
\newline\textit{(Sarcasm)}
& Brainstorm a list of commonly used sarcasm devices in Arabic tweets about the
domain: \{topic domain\}.
\newline\newline
Please adhere to the following guidelines:
\newline
- Sarcasm devices are coarse-grained rhetorical mechanisms, not specific words
or phrases.
\newline
- They describe the way sarcasm is expressed, e.g. verbal irony, hyperbole,
mock praise.
\newline
- Always include `direct\_statement' as the non-sarcastic baseline device.
\newline\newline
Your output should be a Python list of strings, with each element being a
device name. Output ONLY the Python list, nothing else. \\
\hline

Linguistic Device
\newline\textit{(Sentiment)}
& Brainstorm a list of commonly used sentiment categories in Arabic tweets
about the domain: \{topic domain\}.
\newline\newline
Please adhere to the following guidelines:
\newline
- Sentiment categories are coarse-grained expressive modes, not specific words
or phrases.
\newline
- They describe the way a polarity is expressed, e.g. frustration, celebration,
factual reporting.
\newline
- Provide categories for each of the three polarities: positive, neutral, and
negative.
\newline\newline
Your output should be a Python dict with keys ``positive'', ``neutral'', and
``negative'', each mapping to a list of category names.
Output ONLY the Python dict, nothing else. \\
\hline

Linguistic Marker
& Brainstorm a list of Arabic words or phrases that signal \{linguistic device\}
in tweets about \{topic domain\}.
\newline\newline
Please adhere to the following guidelines:
\newline
- Markers are fine-grained and concrete surface cues, not broad descriptions.
\newline
- A marker is text a native speaker would immediately associate with the given
device, dialect, or polarity.
\newline
- Include lexical items, multiword phrases, and, for dialects, characteristic
morphological and phonological forms.
\newline
- Provide diverse markers spanning formal, colloquial, and mixed register.
\newline
- Write each marker in Arabic script.
\newline\newline
Your output should be a Python list of Arabic strings.
Output ONLY the Python list, nothing else. \\
\hline
\end{tabular}
\caption{Brainstorming prompts used in BARRAC. The linguistic device prompt is
task-specific; the tweet topic and linguistic marker prompts are shared across
tasks, with the device slot instantiated per task. For dialect, the device set
is fixed a priori by the label space (EGY, GLF, LEV, MGH, MSA)}
\label{tab:barrac-brainstorm}
\end{table*}

\begin{table}[t]
\centering
\footnotesize
\begin{tabular}{p{1.4cm}|p{5.6cm}}
\hline
\textbf{Task} & \textbf{Task Instruction / Label Space} \\
\hline
Sarcasm
& identifying whether the tweet is sarcastic or not sarcastic.
\newline
\textit{Label space:} sarcastic, not sarcastic \\
\hline
Dialect
& identifying which Arabic dialect the tweet is written in.
\newline
\textit{Label space:} EGY, GLF, LEV, MGH, MSA \\
\hline
Sentiment
& identifying the sentiment of the tweet.
\newline
\textit{Label space:} positive, neutral, negative \\
\hline
\end{tabular}
\caption{Task instructions and label spaces substituted into the
\{task instruction\} and \{label space\} slots of all BARRAC prompts.}
\label{tab:barrac-instructions}
\end{table}

\begin{table*}[t]
\centering
\footnotesize
\begin{tabular}{p{2.2cm}|p{12.5cm}}
\hline
\textbf{Method} & \textbf{Prompt} \\
\hline

Attribute
\newline Prompting
& Write a tweet for the Arabic Twitter: \{tweet topic\} Label the tweet by
\{task instruction\}.
\newline\newline
Requirements:
\newline
- Keep a consistent style and annotation standard with the examples.
\newline
- Convey {linguistic device} in the tweet, using the {marker type} 
  '{linguistic marker}' as a signal.
\newline
- Keep a consistent tweet style with examples.
\newline
- Express \{expression pattern\} across the tweet.
\newline\newline
Here are some examples:
\newline
Tweet: \{tweet\}
\newline
Label: \{label\}
\newline
\ldots\ \ldots
\newline\newline
Tweet: <Arabic tweet text>
\newline
Label: <\{label space\}> \\
\hline
\end{tabular}
\caption{Prompt template for attribute prompting in BARRAC.
\{task instruction\} and \{label space\} are instantiated per task
(Table~\ref{tab:barrac-instructions}). \{expression pattern\} takes the values
consistent, mixed, or implicit for sarcasm and sentiment, and formal,
colloquial, or mixed for dialect.}
\label{tab:barrac-attribute}
\end{table*}

\begin{table*}[t]
\centering
\footnotesize
\begin{tabular}{p{2.2cm}|p{12.5cm}}
\hline
\textbf{Method} & \textbf{Prompt} \\
\hline

\multirow{20}{2.2cm}{Sample \newline Combination}
& Given 2 Arabic Twitter example tweets with the labels, please combine them to
generate \{K\} diverse tweets. Label each tweet by \{task instruction\}.
\newline\newline
Requirements:
\newline
- Keep a consistent style and annotation standard with the examples.
\newline
- Maintain the same format as the examples.
\newline
- Combine the aspects and meanings of both examples in every generated tweet.
\newline
- If both examples share the same label, all \{K\} generated tweets must carry
that label; otherwise, generate an equal number of tweets for each label.
\newline\newline
Examples:
\newline
Tweet: \{tweet\}
\newline
Label: \{label\}
\newline
Tweet: \{tweet'\}
\newline
Label: \{label'\}
\newline\newline
\{K\} Diverse Combined Tweets with Labels:
\newline
1. Tweet: \\
% \cdashline{2-2}

& Given an Arabic Twitter example tweet with the label, please paraphrase it to
generate \{K\} diverse tweets. Label each tweet by \{task instruction\}.
\newline\newline
Requirements:
\newline
- Keep a consistent style and annotation standard with the example.
\newline
- Maintain the same format as the example.
\newline
- The meaning of the example tweet should be unchanged.
\newline
- All generated tweets must carry the same label as the example.
\newline\newline
Example:
\newline
Tweet: \{tweet\}
\newline
Label: \{label\}
\newline\newline
\{K\} Diverse Paraphrased Tweets with Labels:
\newline
1. Tweet: \\
\hline

\multirow{14}{2.2cm}{Selective \newline Reconstruction}
& Given a partially masked Arabic Twitter tweet, please fill and reconstruct it
to generate \{K\} diverse tweets following the example. Label each tweet by
\{task instruction\}.
\newline\newline
Masked tweet: \{masked tweet\}
\newline\newline
Requirements:
\newline
- Keep a consistent style and annotation standard with the example.
\newline
- Maintain the same format as the example.
\newline
- The unmasked part of the tweet should be unchanged.
\newline\newline
Example:
\newline
Tweet: \{tweet\}
\newline
Label: \{label\}
\newline\newline
\{K\} Diverse Reconstructed Tweets with Labels:
\newline
1. Tweet: \\
\hline
\end{tabular}
\caption{Prompts for instance-driven data synthesis in BARRAC. The reconstruction prompt is applied to both
masking strategies, differing only in the masked tweet supplied.}
\label{tab:barrac-instance}
\end{table*}

\begin{table*}[t]
\centering
\footnotesize
\begin{tabular}{p{1.9cm}|p{12.8cm}}
\hline
\textbf{Method} & \textbf{Input Prompt / Output} \\
\hline

\multirow{30}{1.9cm}{Attribute \newline Prompting \newline\newline \textit{(Sarcasm)}}
& Write a tweet for the Arabic Twitter: \textcolor{blue}{a government
announcement about electricity price increases} Label the tweet by
\textcolor{blue}{identifying whether the tweet is sarcastic or not sarcastic}.
\newline\newline
Requirements:
\newline
- Keep a consistent style and annotation standard with the examples.
\newline
- Convey \textcolor{blue}{mock\_praise} in the tweet, using the
\textcolor{blue}{sarcasm marker} `\textarabic{يا سلام}'
(\textit{``how wonderful''}) as a signal.
\newline
- Keep a consistent tweet style with examples.
\newline
- Express \textcolor{blue}{consistent} across the tweet.
\newline\newline
% Here are some examples:
% \newline\newline
% Tweet: \textarabic{ارتفعت أسعار الوقود مجدداً هذا الشهر وفق بيان رسمي}
% \newline
% \textit{Fuel prices rose again this month according to an official statement}
% \newline
% Label: not sarcastic
% \newline\newline
% Tweet: \textarabic{برافو على الخدمات الرائعة اللي ما شفناها}
% \newline
% \textit{Bravo on the wonderful services we never saw}
% \newline
% Label: sarcastic
% \newline\newline
% Tweet: \textarabic{تعلن الوزارة عن جدول الانقطاعات المقرر للأسبوع القادم}
% \newline
% \textit{The ministry announces the outage schedule for next week}
% \newline
% Label: not sarcastic
% \newline\newline
% Tweet: \textarabic{يا سلام على التوقيت المثالي، رفع الأسعار في عز الصيف}
% \newline
% \textit{How wonderful, perfect timing raising prices in the
% peak of summer}
% \newline
% Label: sarcastic
\newline
% \ldots\ \ldots

Tweet: <Arabic tweet text>
\newline
Label: <sarcastic or not sarcastic> \\
\\
% \cdashline{2-2}
\\

& 
\par
\ldots\ \ldots

Tweet: \textarabic{يا سلام على القرار الحكيم، رفعوا الكهرباء وإحنا بنشكرهم}
\newline
\textit{How wonderful, a wise decision, they raised electricity
prices and we thank them}
\newline
Label: sarcastic \\

\hline
\end{tabular}
\caption{Example of key-point-driven synthetic data from LLMs in BARRAC.
Slots instantiated from the sampled attribute tuple and the task instruction
are shown in \textcolor{blue}{blue}; English glosses are given in italics and
do not form part of the prompt. The output below the dashed line is the model
response.}
\label{tab:barrac-example-kpd}
\end{table*}

\begin{table*}[t]
\centering
\scriptsize
\begin{tabular}{p{1.9cm}|p{12.8cm}}
\hline
\textbf{Method} & \textbf{Input Prompt / Output} \\
\hline

\multirow{28}{1.9cm}{Sample \newline Combination \newline\newline \textit{(Sentiment)}}
& Given 2 Arabic Twitter example tweets with the labels, please combine them to
generate \textcolor{blue}{4} diverse tweets. Label each tweet by
\textcolor{blue}{identifying the sentiment of the tweet}.
\newline\newline
Requirements:
\newline
- Keep a consistent style and annotation standard with the examples.
\newline
- Maintain the same format as the examples.
\newline
- Combine the aspects and meanings of both examples in every generated tweet.
\newline
- If both examples share the same label, all 4 generated tweets must carry
that label; otherwise, generate an equal number of tweets for each label.
\newline\newline
Examples:
\newline
Tweet: \textarabic{أسعار الخضار ارتفعت بشكل جنوني هذا الأسبوع}
\newline
\textit{Vegetable prices rose insanely this week}
\newline
Label: negative
\newline
Tweet: \textarabic{الزحمة في الطريق خانقة من الصبح}
\newline
\textit{The traffic on the road has been suffocating since morning}
\newline
Label: negative
\newline\newline
4 Diverse Combined Tweets with Labels:
\newline
1. Tweet: \\
\cline{2-2}

& 
\par
\ldots\ \ldots

\textarabic{وقفت ساعة في الزحمة عشان أشتري خضار بأسعار جنونية}
\newline
\textit{I stood an hour in traffic just to buy vegetables at insane prices}
\newline
Label: negative
\newline\newline
2. Tweet: \textarabic{الزحمة خانقة والأسعار في السوق ما تطاق، يوم متعب}
\newline
\phantom{2. }\textit{The traffic is suffocating and market prices are
unbearable, an exhausting day}
\newline
\phantom{2. }Label: negative
\newline\newline
3. Tweet: \textarabic{من زحمة الصبح لغلاء الخضار، كل شيء صار أصعب}
\newline
\phantom{3. }\textit{From the morning traffic to the cost of vegetables,
everything has become harder}
\newline
\phantom{3. }Label: negative
\newline\newline
4. Tweet: \textarabic{أسعار الخضار جنونية والطريق خانق، الوضع صعب هالأسبوع}
\newline
\phantom{4. }\textit{Vegetable prices are insane and the road is choked,
the situation is hard this week}
\newline
\phantom{4. }Label: negative \\
\hline

\multirow{26}{1.9cm}{Selective \newline Reconstruction \newline\newline \textit{(Dialect)}}
& Given a partially masked Arabic Twitter tweet, please fill and reconstruct it
to generate \textcolor{blue}{4} diverse tweets following the example. Label
each tweet by \textcolor{blue}{identifying which Arabic dialect the tweet is
written in}.
\newline\newline
Masked tweet: \textarabic{إيه ده} <mask> <mask> <mask> \textarabic{من الصبح}
\newline
\textit{What is this <mask> <mask> <mask> since morning}
\newline\newline
Requirements:
\newline
- Keep a consistent style and annotation standard with the example.
\newline
- Maintain the same format as the example.
\newline
- The unmasked part of the tweet should be unchanged.
\newline\newline
Example:
\newline
Tweet: \textarabic{إيه ده كله زحمة كده من الصبح}
\newline
\textit{What is all this traffic since morning}
\newline
Label: EGY
\newline\newline
4 Diverse Reconstructed Tweets with Labels:
\newline
1. Tweet: \\
\cline{2-2}

& 

\par
\ldots\ \ldots

\textarabic{إيه ده كله زحمة كده من الصبح}
\newline
\textit{What is all this traffic since morning}
\newline
Label: EGY
\newline\newline
2. Tweet: \textarabic{إيه ده الجو حر أوي من الصبح}
\newline
\phantom{2. }\textit{What is this, the weather has been so hot since morning}
\newline
\phantom{2. }Label: EGY
\newline\newline
3. Tweet: \textarabic{إيه ده الطابور مش بيخلص من الصبح}
\newline
\phantom{3. }\textit{What is this, the queue hasn't ended since morning}
\newline
\phantom{3. }Label: EGY
\newline\newline
4. Tweet: \textarabic{إيه ده الشغل كتر كده من الصبح}
\newline
\phantom{4. }\textit{What is this, the work has piled up like this since
morning}
\newline
\phantom{4. }Label: EGY \\
\hline
\end{tabular}
\caption{Examples of instance-driven synthetic data from LLMs in BARRAC.
Slots instantiated per task are shown in \textcolor{blue}{blue}; English
glosses are given in italics and do not form part of the prompt. The output
below each dashed line is the model response for $K=4$. The output below the dashed line is the model
response.}
\label{tab:barrac-example-idd}
\end{table*}

\end{document}